\documentclass[10pt,twocolumn,letterpaper]{article}

\usepackage{cvpr}
\usepackage{times}
\usepackage{epsfig}
\usepackage{graphicx}
\usepackage{amsmath}
\usepackage{amssymb}
\usepackage{float}
\usepackage{multirow}
\usepackage{soul}
\usepackage{subcaption}
\usepackage{tensor}
\usepackage[dvipsnames]{xcolor}
\usepackage{bbold} 
\usepackage{mathtools} 

\def\ours{\texttt{\textbf{TCL}}\xspace}

\usepackage[pagebackref=true,breaklinks=true,letterpaper=true,colorlinks,bookmarks=false]{hyperref}

\cvprfinalcopy 

\ifcvprfinal\fi
\begin{document}

\title{Semi-Supervised Action Recognition with Temporal Contrastive Learning}
\author{Ankit Singh$^{1}$\thanks{The first two authors contributed equally.} \ \ \ \ Omprakash Chakraborty$^{2}$\textsuperscript{*} \ \ \ \ Ashutosh Varshney$^{2}$ \ \ \ \ Rameswar Panda$^{3}$ \\ Rogerio Feris$^{3}$ \ \ \ \ Kate Saenko$^{3,4}$ \ \ \ \ Abir Das$^{2}$ \\
$^{1}$ IIT Madras, $^{2}$ IIT Kharagpur, $^{3}$ MIT-IBM Watson AI Lab, $^{4}$ Boston University
}
\maketitle
\thispagestyle{empty}

\begin{abstract}
Learning to recognize actions from only a handful of labeled videos is a challenging problem due to the scarcity of tediously collected activity labels. We approach this problem by learning a two-pathway temporal contrastive model using unlabeled videos at two different speeds leveraging the fact that changing video speed does not change an action. Specifically, we propose to maximize the similarity between encoded representations of the same video at two different speeds as well as minimize the similarity between different videos played at different speeds. This way we use the rich supervisory information in terms of `time' that is present in otherwise unsupervised pool of videos. With this simple yet effective strategy of manipulating video playback rates, we considerably outperform video extensions of sophisticated state-of-the-art semi-supervised image recognition methods across multiple diverse benchmark datasets and network architectures. Interestingly, our proposed approach benefits from out-of-domain unlabeled videos showing generalization and robustness. We also perform rigorous ablations and analysis to validate our approach.
Project page:  \href{https://cvir.github.io/TCL/}{https://cvir.github.io/TCL/}.
\end{abstract}

\section{Introduction}
\label{sec:introduction}
Supervised deep learning approaches have shown remarkable progress in video action recognition~\cite{Carreira2017I3D, Fan2019Tam, Feichtenhofer2020X3D, Feichtenhofer2019Slowfast, Lin2019Tsm, Tran2015C3D}.
However, being supervised, these models are critically dependent on large datasets requiring tedious human annotation effort.
This motivates us to look beyond the supervised setting as supervised methods alone may not be enough to deal with the volume of information contained in videos.
Semi-supervised learning approaches use structural invariance between different views of the same data as a source of supervision for learning useful representations.
In recent times, semi-supervised representation learning models \cite{Chen2020Simple, Henaff2019Data, Misra2020Self, Oord2018Representation} have performed very well even surpassing its supervised counterparts in case of images~\cite{Goyal2019Scaling, Sohn2020Fixmatch}.

\begin{figure}[t]
  \centering
  \includegraphics[width=\columnwidth]{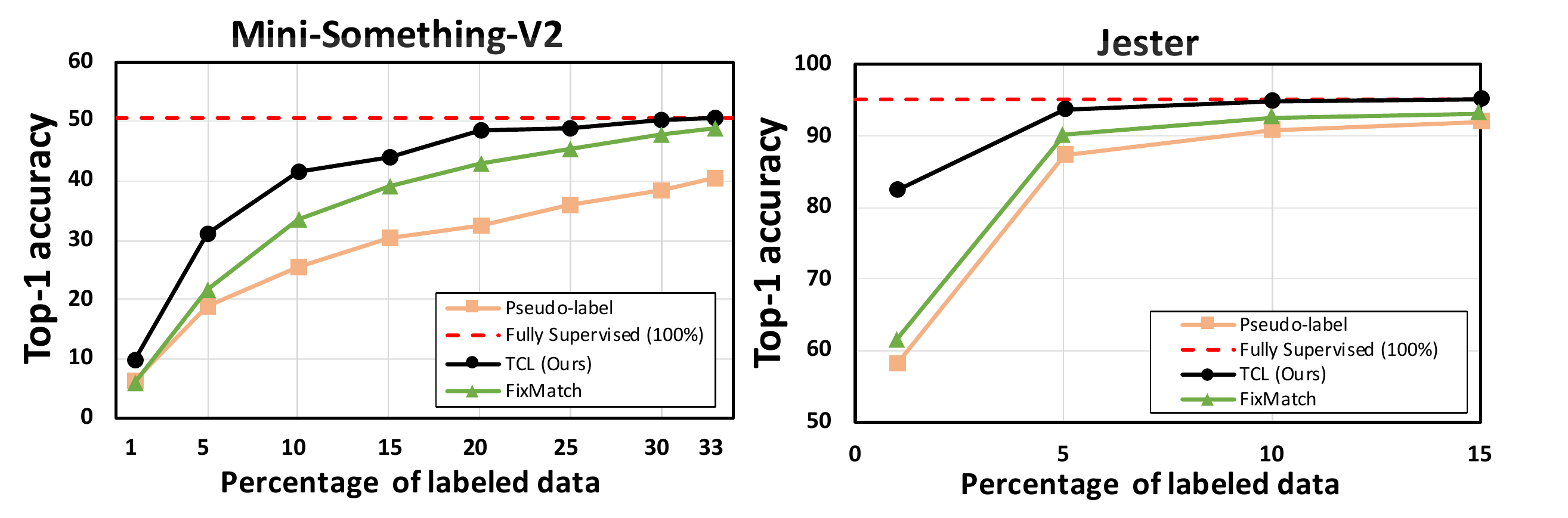} \vspace{-6mm}
  \caption{\small \textbf{Comparison of top-1 accuracy for \ours (Ours) with Pseudo-Label~\cite{Lee2013Pseudo} and FixMatch~\cite{Sohn2020Fixmatch} baselines trained with different percentages of labeled training data.} We evaluate the efficacy of the approaches in terms of the least proportion of labeled data required to surpass the fully supervised ~\cite{Lin2019Tsm} performance (shown with the {\color{red}{red}} dotted line). With only 33\% and 15\% of labeled data, our proposed \ours framework surpasses the supervised approaches in Mini-Something-V2~\cite{Goyal2017Something} and Jester~\cite{materzynska2019jester} datasets respectively. 
  The two other compared methods fail to reach the accuracy of the fully supervised approach with such small amount of labeled data. (Best viewed in color.)
  }
  \label{fig:fully-sup-comparison} \vspace{-4mm}
\end{figure}

Notwithstanding their potential, semi-supervised video action recognition has received very little attention.
Trivially extending the image domain approaches to videos without considering the rich temporal information may not quite bridge the performance gap between the semi and the fully supervised learning.
But, in videos, we have another source of supervision: \emph{time}.
We all know that an action recognizer is good if it can recognize actions irrespective of whether the actions are performed slowly or quickly.
Recently, supervised action recognition has benefited a lot by using differently paced versions of the same video during training~\cite{Feichtenhofer2019Slowfast, xiao2020Audiovisual}.
Motivated by the success of using slow and fast versions of videos for supervised action recognition as well as by the success of the contrastive learning frameworks~\cite{Han2020Memory, Qian2020Spatiotemporal}, we propose \emph{Temporal Contrastive Learning} (\ours) for semi-supervised action recognition in videos where consistent features representing both slow and fast versions of the same videos are learned.

Starting with a model trained with limited labeled data,
we present a two-pathway model that processes unlabeled videos at two different speeds and finds their representations.
Though played at two different speeds, the videos share the same semantics.
Thus, similarity between these representations are maximized.
Likewise, the similarity between the representations of different videos are minimized.
We achieve this by minimizing a modified NT-Xent contrastive loss~\cite{Chen2020Simple, Oord2018Representation} between these videos with different playback rates.
While minimizing a contrastive loss helps to produce better visual representations by learning to be invariant to different views of the data, it ignores information shared among samples of same action class as the loss treats each video individually.
To this end, we propose a new perspective of contrastive loss between neighborhoods.
Neighborhoods are compact groups of unlabeled videos with high class consistency.
In absence of ground-truth labels, groups are formed by clustering videos with same pseudo-labels and are represented by averaging the representations of the constituent videos.
Contrastive objective between groups formed off the two paths explores the underlying class concept that traditional NT-Xent loss among individual video instances does not take into account.
We term the contrastive loss considering only individual instances as the \emph{instance-contrastive loss} and the same between the groups as the \emph{group-contrastive loss} respectively.

We perform extensive experiments on four standard datasets and demonstrate that
\ours achieves superior performance over extended baselines of state-of-the-art image domain semi-supervised approaches.
Figure~\ref{fig:fully-sup-comparison} shows comparison of \ours with Pseudo-Label~\cite{Lee2013Pseudo} and FixMatch~\cite{Sohn2020Fixmatch} trained using different percentages of labeled training data.
Using the same backbone network (ResNet-18), \ours needs only \textbf{33\%} and \textbf{15\%} of labeled data in Mini-Something-V2~\cite{chen2020deep} and Jester~\cite{materzynska2019jester} respectively to reach the performance of the fully supervised approach~\cite{Lin2019Tsm} that uses 100\% labeled data. On the other hand, the two compared methods fail to reach the accuracy of the fully supervised approach with such small amount of labeled data. 
Likewise, we observe as good as 8.14\% and 4.63\% absolute improvement in recognition performance over the next best approach, FixMatch~\cite{Sohn2020Fixmatch} using only 5\% labeled data in Mini-Something-V2~\cite{chen2020deep} and Kinetics-400~\cite{kay2017kinetics} datasets respectively.
In a new realistic setting, we argue that unlabeled videos may come from a related but different domain than that of the labeled data.
For instance, given a small set of labeled videos from a third person view, our approach is shown to benefit from using only first person unlabeled videos on Charades-Ego~\cite{sigurdsson2018charades} dataset, demonstrating the robustness to domain shift in the unlabeled set.
To summarize, our key contributions include:
\begin{itemize}
\setlength{\itemsep}{-0.2pt}
    \item First of all, we treat the time axis in unlabeled videos specially, by processing them at two different speeds and propose a two-pathway temporal contrastive semi-supervised action recognition framework.
    \item Next, we identify that directly employing contrastive objective instance-wise on video representations learned with different frame-rates may miss crucial information shared across samples of same inherent class.
    A novel group-contrastive loss is pioneered to couple discriminative motion representation with pace-invariance that significantly improves semi-supervised action recognition performance.
    \item We demonstrate through experimental results on four datasets, \ours's superiority over extended baselines of successful image-domain semi-supervised approaches.
    The versatility and robustness of our approach in case of training with unlabeled videos from a different domain is shown along with in-depth ablation analysis pinpointing the role of the different components.
\end{itemize}

\section{Related Work}
\label{sec:relatedwork}
\noindent\textbf{Action Recognition.} 
Action recognition is a challenging problem with great application potential.
Conventional approaches based on deep neural networks are mostly built over a two-stream CNN based framework~\cite{Simonyan2014Two}, one to process a single RGB frame and the other for optical flow input to analyze the spatial and temporal information respectively.
Many variants of 3D-CNNs such as C3D~\cite{Tran2015C3D}, I3D~\cite{Carreira2017I3D} and ResNet3D~\cite{hara2017learning}, that use 3D convolutions to model space and time jointly, have also been introduced for action recognition. SlowFast network~\cite{Feichtenhofer2019Slowfast} employs two pathways for recognizing actions by processing a video at both slow and fast frame rates.
Recent works also utilize 2D-CNNs for efficient video classification by using different temporal aggregation modules such as temporal averaging in TSN~\cite{Wang2016Temporal}, bag of features in TRN~\cite{zhou2018temporal}, channel shifting in TSM~\cite{Lin2019Tsm}, depthwise convolutions in TAM~\cite{Fan2019Tam}. Despite promising results on common benchmarks, these models are critically dependent on large datasets that require careful and tedious human annotation effort. In contrast, we propose a simple yet effective temporal contrastive learning framework for semi-supervised action recognition that alleviates the data annotation limitation of supervised methods.

\begin{figure*}[ht]
  \centering
   \includegraphics[width=\textwidth]{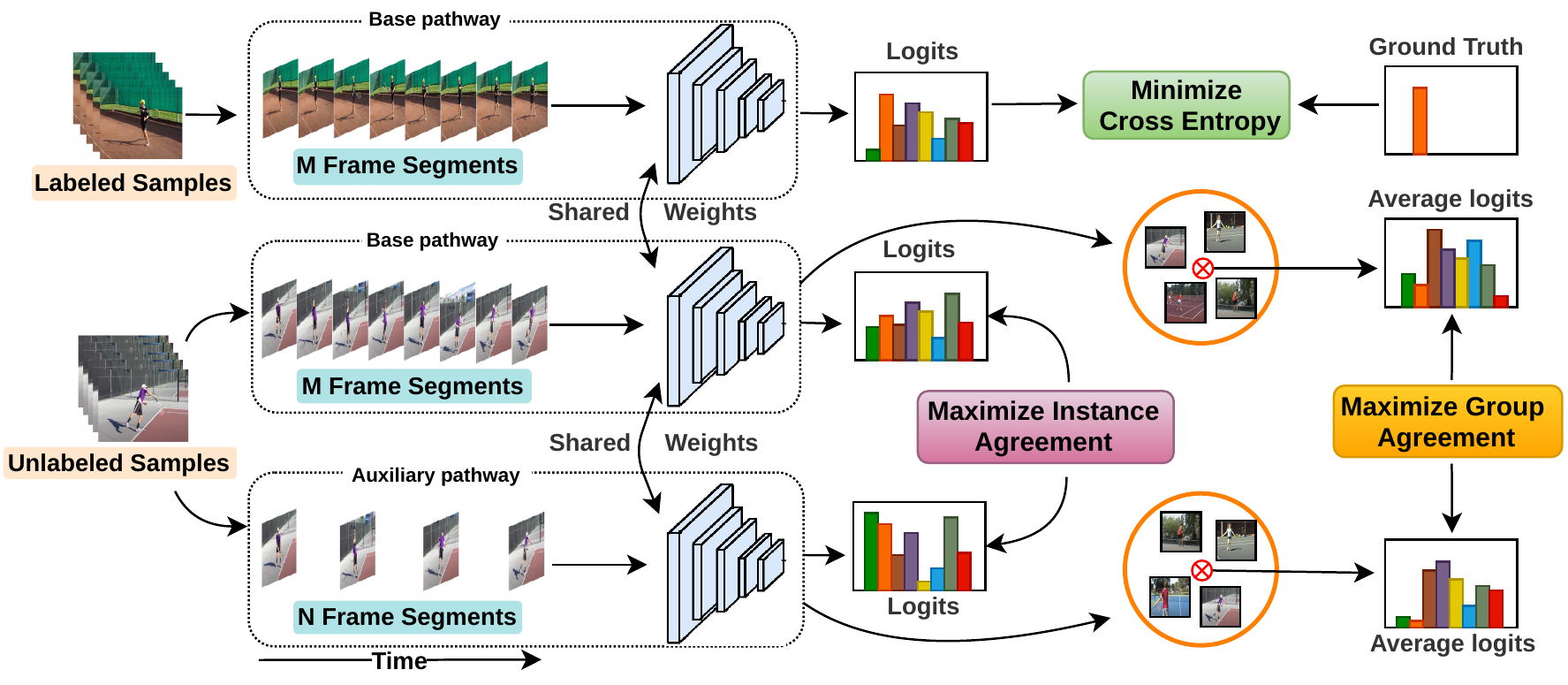} \vspace{-7mm}
  \caption{\small \textbf{Illustration of our Temporal Contrastive Learning (\ours) Framework.} Our approach consists of \emph{base} and \emph{auxiliary} pathways that share the same weights. Base pathway accepts video frames sampled at a higher rate while the auxiliary pathway takes in frames at a lower framerate. At first, the base network is trained using limited labeled data. Subsequently, the auxiliary pathway comes into picture for the unlabeled samples by encouraging video representations to match in both pathways in absence of labels. This is done by maximizing agreement between the outputs of the two pathways for a video while minimizing the same for different videos. In addition, originally unlabeled videos with high semantic similarity are grouped by pseudo-labels assigned to them. To exploit the high consistency and compactness of group members, the average representations of groups with same pseudo-label in different pathways are made similar while that between the varying groups are made maximally different.
 Two separate contrastive losses (ref Sections~\ref{subsubsec:instance-contrastive} and \ref{subsubsec:group-consistency}) are used for this purpose. Given a video at test time, only the base network is used to recognize the action. (Best viewed in color.)}
  \label{fig:overview-figure} \vspace{-3mm}
\end{figure*}

\vspace{1mm}
\noindent\textbf{Semi-Supervised Learning.} 
Semi-supervised learning (SSL) has been studied from multiple aspects (see reviews~\cite{chapelle2009semi}).
Various strategies have been explored \textit{e.g.}, generative models~\cite{odena2016semi,rasmus2015semi}, self-training using pseudo-labels~\cite{arazo2020pseudo, grandvalet2005semi, Lee2013Pseudo} and consistency regularization~\cite{Bachman2014Learning, berthelot2019remixmatch, Berthelot2019Mixmatch, Laine2016Temporal, miyato2018virtual, Tarvainen2017Mean, xie2019unsupervised}.
Leveraging self-supervised learning like rotation prediction~\cite{Gidaris2018Unsupervised} and image transformations~\cite{dosovitskiy2014discriminative} is also another recent trend for SSL~\cite{Zhai2019S4L}.
While there has been tremendous progress in semi-supervised image classification, SSL for action recognition is still a novel and rarely addressed problem. Iosifidis \textit{et al.}~\cite{iosifidis2014semi}, first utilize traditional Action Bank for action representation and then uses a variant of extreme learning machine for semi-supervised classification.
The work most related to ours is~\cite{Sohn2020Fixmatch} which first generates confident one-hot labels for unlabelled images and then trains the model to be consistent across different forms of image augmentations.
While this has recently achieved great success, the data augmentations for generating different transformations are limited to transformations in the image space and fail to leverage the temporal information present in videos.
We differ from~\cite{Sohn2020Fixmatch} as we propose a temporal contrastive learning framework for semi-supervised action recognition by modeling temporal aspects using two pathways at different speeds instead of augmenting images.
We further propose a group-wise contrastive loss in addition to instance-wise contrastive loss for learning discriminative features for action recognition.   

\noindent\textbf{Contrastive Learning.}
Contrastive learning~\cite{Chen2020Simple, Chen2020Big, Fotedar2020Extreme, he2020momentum, hjelm2018learning, korbar2018cooperative, Misra2020Self, Oord2018Representation, Wu2018Unsupervised} is becoming increasingly attractive due to its great potential to leverage large amount of unlabeled data.
The essence of contrastive learning lie in maximizing the similarity of representations among positive samples while encouraging discrimination for negative samples.
Some recent works have also utilized contrastive learning~\cite{Gordon2020Watching, Han2019Video, Qian2020Spatiotemporal, Sermanet2018Time, yang2020video} for self-supervised video representation learning.
Spatio-temporal contrastive learning using different augmentations for learning video features is presented in~\cite{Qian2020Spatiotemporal}. 
Speed of a video is also investigated for self-supervised learning~\cite{benaim2020speednet, wang2020self, yao2020video} unlike the problem we consider in this paper.
While our approach is inspired by these, we focus on semi-supervised action recognition in videos, where our goal is to learn consistent features representing two different frame rates of the unlabeled videos.

\section{Methodology}
\label{sec:methodology}
In this section, we present our novel semi-supervised approach to efficiently learn video representations. First we briefly discuss the problem description and then describe our framework and its components in detail.

\subsection{Problem Setup}
Our aim is to address semi-supervised action recognition where only a small set of videos ($\mathcal{D}_l$) has labels, but a large number of unlabeled videos ($\mathcal{D}_u$) are assumed to be present alongside.
The set $\mathcal{D}_l \triangleq \{V^i,y^i\}_{i=1}^{N_l}$ comprises of $N_l$ videos where the $i^{th}$ video and the corresponding activity label is denoted by $V^i$ and $y^i$ respectively. For a dataset of videos with $C$ different activities, $y^i$ is often assumed to be an element of the label set $Y = \{1, 2, \cdots, C\}$.
Similarly, the unlabeled set $\mathcal{D}_u \triangleq \{U^i\}_{i=1}^{N_u}$ comprises of $N_u (\gg\!\! N_l)$ videos without any associated labels.
We use the unlabeled videos at two different frame rates and refer to them as fast and slow videos.
The fast version of the video $U^i$ is represented as a collection of $M$ frames \textit{i.e.}, $U_f^i=\{F_{f,1}^i, F_{f,2}^i, \cdots, F_{f,M}^i\}$.
Likewise, the slow version of the same is represented as $U_s^i=\{F_{s,1}^i, F_{s,2}^i, \cdots, F_{s,N}^i\}$ where $N < M$. The frames are sampled from the video following Wang \textit{et. al}~\cite{Wang2016Temporal} where a random frame is sampled uniformly from consecutive non-overlapping segments.

\subsection{Temporal Contrastive Learning}
\label{subsec:TCL}
As shown in Figure~\ref{fig:overview-figure}, our `Temporal Contrastive Learning (\ours)' framework processes the input videos in two pathways, namely \textit{base} and \textit{auxiliary} pathways.
The fast version of the videos are processed by \textit{base} while the slow versions are processed by the \textit{auxiliary} pathway.
Both pathways share same neural backbone (denoted by $g(.)$).
Different stages of training in \ours framework are described next.

\subsubsection{Supervised Training Stage} 
\label{subsubsec:sup}

The neural network backbone is initially trained using only the small labeled data $\mathcal{D}_l$ by passing it through the base branch. Depending on whether the backbone involves 2D~\cite{Lin2019Tsm, Wang2016Temporal} or 3D convolution~\cite{Carreira2017I3D, hara2017learning} operations, the representation ($g(V^i)$) of the video $V^i$ used in our framework is average of the frame logits or the logits from the 3D backbone respectively.
We minimize the standard supervised cross-entropy loss ($\mathcal{L}_{sup}$) on the labeled data as follows.

\vspace{-2mm}
\begin{equation}
\label{eq:sup_loss}
\mathcal{L}_{sup} = -\sum\limits_{c=1}^{C}(y^i)_c \log(g(V^i))_c
\vspace{-2mm}
\end{equation}

\subsubsection{Instance-Contrastive Loss} 
\label{subsubsec:instance-contrastive}

Equipped with an initial backbone trained with limited supervision, our goal is to learn a model that can use a large pool of unlabeled videos for better activity understanding.
To this end, we use temporal co-occurrence of unlabeled activities at multiple speeds as a proxy task and enforce this with a pairwise contrastive loss.
Specifically, we adjust the frame sampling rate to get videos with different speeds.
Let us consider a minibatch with $B$ unlabeled videos.
The model is then trained to match the representation $g(U_f^i)$ of the comparatively faster version of the video ($U^i$) with that ($g(U_s^i)$) of the slower version.
$g(U_f^i)$ and $g(U_s^i)$ forms the positive pair.
For rest of $B-1$ videos, $g(U_f^i)$ and $g(U_p^k)$ form negative pairs where representation of $k^{th}$ video can come from either of the pathways (\textit{i.e.}, $p\in\{f, s\}$).
As different videos forming the negative pairs, have different content, the representation of different videos in either of the pathways are pushed apart.
This is achieved by employing a contrastive loss ($\mathcal{L}_{ic}$) as follows.
\small
\begin{equation}
\label{eq:instance}
\hspace{-4mm}
\mathcal{L}_{ic}(U_f^i, U_s^i)\!\!= \!-\!\!\log\!\frac{h\big(g(U_f^i), \!g(U_s^i)\big)}{\!\!h\big(g(U_f^i),\!g(U_s^i)\!\big)\!+\!\!\!\!\!\!\!\sum\limits_{\substack{k=1\\p\in\{s,f\}}}^{B}\!\!\!\!\!\!\mathbb{1}_{\{k \neq i\}}h\big(g(U_f^i),\!g(U_p^k)\!\big)}
\end{equation}
\normalsize
where $h(\mathbf{u},\mathbf{v}) = \exp\big(\frac{\mathbf{u}^\top \mathbf{v}}{||\mathbf{u}||_2||\mathbf{v||_2}}/\tau\big)$
is the exponential of cosine similarity measure and $\tau$ is the temperature hyperparameter.
The final instance-contrastive loss is computed for all positive pairs, i.e., both $(U_f^i, U_s^i)$ and $(U_s^i, U_f^i)$ across minibatch.
The loss function encourages to decrease the similarity not only between different videos in two pathways but also between different videos across both of them.

\begin{figure}[t!]
  \centering
   \includegraphics[width=\columnwidth]{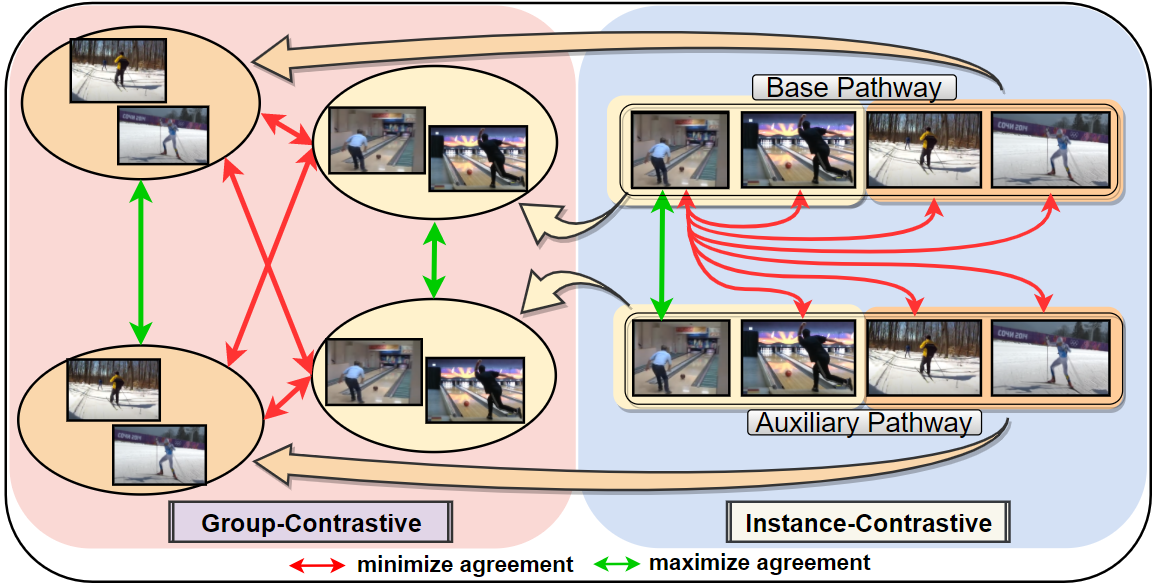} \vspace{-6mm}
  \caption{\small \textbf{Advantage of group-contrastive loss over instance-contrastive loss.}
  A contrastive objective between instances may try to push different instances of same action apart (right), while forming groups of videos with same activity class avoids such inadvertent competition (left). In absence of true labels, such grouping is done by the predicted pseudo-labels. (Best viewed in color.)
  }
  \label{fig:gc_ic} \vspace{-2mm}
\end{figure}

\subsubsection{Group-Contrastive Loss}
\label{subsubsec:group-consistency}

Directly applying contrastive loss between different video instances in absence of class-labels does not take the high level action semantics into account.
As illustrated in Figure~\ref{fig:gc_ic}, such a strategy can inadvertently learn different representations for videos containing same actions.
We employ contrastive loss among groups of videos with similar actions where relations within the neighborhood of different videos are explored.
Specifically, each unlabeled video $U^i$ in each of the two pathways are assigned pseudo-labels that correspond to the class having the maximum activation.
Let $\hat{y}_f^i$ and $\hat{y}_s^i$ denote the pseudo-labels of the video $U^i$ in the fast and the slow pathways respectively.
Videos having the same pseudo-label in a minibatch form a group in each pathway and the average of the representations of constituent videos provides the representation of the group as shown below.
\vspace{-2mm}
\small
\begin{equation}
\label{eq:group_average}
R_p^l = \frac{\sum\limits_{i=1}^{B} \mathbb{1}_{\{\hat{y}_p^i=l\}}g(U_p^i)}{T}
\vspace{-2mm}
\end{equation}
\normalsize
where $\mathbb{1}$ is an indicator function that evaluates to 1 for the videos with pseudo-label equal to $l \!\in\! Y$ in each pathway $p \!\in\! \{f, s\}$.
$T$ is the number of such videos in the minibatch.

Considering the high class consistency among two groups with same label in two pathways, we require these groups to give similar representations in the feature space.
Thus, in the group-contrastive objective, all pairs $(R_f^l, R_s^l)$ act as positive pairs while the negative pairs are the pairs $(R_f^l, R_p^m)$ with $p \!\in \!\{f, s\}$ and $m \!\in\! Y\!\setminus\! l$ such that the constituent groups are different in either of the pathways.
The group-contrastive loss involving these pairs is,

\vspace{-2mm}
\small
\begin{equation}
\hspace{-3mm}\mathcal{L}_{gc}(R_{f}^l, \!R_{s}^l) \!= \!-\!\log\!\frac{h(R_f^l, \!R_s^l)}{h(R_f^l, \!R_s^l)+\sum\limits_{\substack{m=1\\p\in\{s,f\}}}^{C}\!\!\mathbb{1}_{\{m \neq l\}}h(R_f^l, \!R_p^m)}
\vspace{-1mm}
\end{equation}
\normalsize

Similar to instance-contrastive loss, group-contrastive loss is also computed for all positive pairs - both $(R_f^l, R_s^l)$ and $(R_s^l, R_f^l)$ across the minibatch.
Overall, the loss function for training our model involving the limited labeled data and the unlabeled data is,
\vspace{-1mm}

\small
\begin{equation}
\label{eq:total_loss}
\mathcal{L} = \mathcal{L}_{sup} + \gamma*\mathcal{L}_{ic} + \beta*\mathcal{L}_{gc}
\end{equation}
\normalsize
where, $\gamma$ and $\beta$ are weights of the instance-contrastive and group-contrastive losses respectively. 

\subsection{\ours with Pretraining and Finetuning}
\label{subsec:finetuning}

Self-supervised pretraining has recently emerged as a promising alternative, which not only avoids huge annotation effort but also is better and robust compared to its supervised counterpart in many visual tasks~\cite{Erhan2010Why, xie2020self, zoph2020rethinking}. Motivated by this, we adopt self-supervised pretraining to initialize our model with very minimal change in the framework.
Specifically, we employ self-supervised pretraining at the beginning by considering the whole of the labeled and the unlabeled data $\mathcal{D}_l \cup \mathcal{D}_u$ as unlabeled data only and using instance-contrastive loss $\mathcal{L}_{ic}$ to encourage consistency between representations learned in the two pathways (ref. Eq.~\ref{eq:instance}).
These weights are then used to initialize the base and the auxiliary pathways before our approach commences for semi-supervised learning of video representations.
For effective utilization of unlabeled data, we also finetune the base pathway with pseudo-labels~\cite{Lee2013Pseudo} generated at the end of our contrastive learning, which greatly enhances the discriminabilty of the features, leading to improvement in recognition performance.
We empirically show that starting with the same amount of labeling, both self-supervised pretraining and finetuning with pseudo-labels (Pretraining$\rightarrow$\ours$\rightarrow$Finetuning) benefits more compared to the same after limited supervised training only.

\section{Experiments}
\label{sec:experiments}

In this section, we conduct extensive experiments to show that our \ours framework outperforms many strong baselines on several benchmarks including one with domain shift. We also perform comprehensive ablation experiments to verify the effectiveness of different components in detail.

\subsection{Experimental Setup}

\noindent\textbf{Datasets.} We evaluate
our approach using four datasets, namely Mini-Something-V2~\cite{chen2020deep}, Jester~\cite{materzynska2019jester}, Kinetics-400~\cite{kay2017kinetics} and Charades-Ego~\cite{sigurdsson2018charades}.
Mini-Something-V2 is a subset of Something-Something V2 dataset~\cite{Goyal2017Something} containing 81K training videos and 12K testing videos across 87 action classes. Jester~\cite{materzynska2019jester} contains 119K videos for training and 15K videos for validation across 27 annotated classes for hand gestures.
Kinetics-400~\cite{kay2017kinetics} is one of the most popular large-scale benchmark for video action recognition.
It consists of 240K videos for training and 20K videos for validation across 400 action categories, with each video lasting 6-10 seconds. 
Charades-Ego~\cite{sigurdsson2018charades} contains 7,860 untrimmed egocentric videos of daily indoors activities recorded from both third and first person views. The dataset contains 68,536 temporal annotations for 157 action classes. We use a subset of the third person videos from Charades-Ego as the labeled data while the first person videos are considered as unlabeled data to show the effectiveness of our approach under domain shift in the unlabeled data. More details about the datasets are given in the appendix.

\begin{table*}[ht!]
\centering
\resizebox{\textwidth}{!}{%
\begin{tabular}{l||ccc||ccc}
\hline
& \multicolumn{3}{c||}{ResNet-18}& \multicolumn{3}{c}{ResNet-50}\\
\cline{2-7}
Approach  & 1\% & 5\% & 10\% &  1\% & 5\% & 10\%\\
\hline
 
 Supervised (8f) & 5.98${\pm 0.68}$ & 17.26${\pm 1.17}$ & 24.67${\pm 0.68}$ & 5.69${\pm 0.51}$& 16.68${\pm 0.25}$ & 25.92${\pm 0.53}$ \\
 Pseudo-Label~\cite{Lee2013Pseudo} \small{(\texttt{ICMLW'13})} & 6.46${\pm 0.32}$ & 18.76${\pm 0.77}$ & 25.67${\pm 0.45}$ & 6.66${\pm 0.89}$ & 18.77${\pm 1.18}$ & 28.85${\pm 0.91}$ \\
Mean Teacher~\cite{Tarvainen2017Mean} \small{(\texttt{NeurIPS'17})} & 7.33${\pm 1.13}$ & 20.23${\pm 1.59}$ & 30.15${\pm 0.42}$ & 6.82${\pm 0.18}$ & 21.80${\pm 1.54}$ & 32.12${\pm 2.37}$  \\
S4L~\cite{Zhai2019S4L} \small{(\texttt{ICCV'19})}  & 7.18${\pm 0.97}$ & 18.58${\pm 1.05}$ & 26.04${\pm 1.89}$ & 6.87${\pm 1.29}$ & 17.73${\pm 0.26}$ & 27.84${\pm 0.75}$\\
MixMatch~\cite{Berthelot2019Mixmatch} \small{(\texttt{NeurIPS'19})} & 7.45${\pm 1.01}$ & 18.63${\pm 0.99}$ & 25.78${\pm 1.01}$ & 6.48${\pm 0.83}$ & 17.77${\pm 0.12}$ & 27.03${\pm 1.66}$ \\ 
FixMatch~\cite{Sohn2020Fixmatch} \small{(\texttt{NeurIPS'20})} & 6.04${\pm 0.44}$ & 21.67${\pm 0.18}$ & 33.38${\pm 1.58}$ & 6.54${\pm 0.71}$& 25.34${\pm 2.03}$ & 37.44${\pm 1.31}$  \\
\hline 
\ours (Ours) & 7.79${\pm 0.57}$ & 29.81${\pm 0.77}$ & 38.61${\pm 0.91}$ & 7.54${\pm 0.32}$ & 27.22${\pm 1.86}$ & 40.70${\pm 0.42}$\\
\ours w/ Finetuning& 8.65${\pm 0.76}$ & 30.55${\pm 1.36}$ & 40.06${\pm 1.14}$ & 8.56${\pm 0.31}$ & 28.84${\pm 1.22}$ & 41.68${\pm 0.56}$ \\
\ours w/ Pretraining \& Finetuning & 9.91${\pm 1.84}$ & 30.97${\pm 0.07}$ & 41.55${\pm 0.47}$ &  9.19${\pm 0.43}$ & 29.85${\pm 1.76}$ & 41.33${\pm 1.07}$ \\ 
\hline
\end{tabular}
} \vspace{-2mm}
\caption{\small \textbf{Performance Comparison in Mini-Something-V2.} Numbers show average Top-1 accuracy values with standard deviations over 3 random trials for different percentages of labeled data. \ours significantly outperforms all the compared methods in both cases.
}
\label{tab:comp_sth_v2_mini} \vspace{-2mm}
\end{table*}
\begin{table*}[ht!]
\centering
\resizebox{\textwidth}{!}{%
\begin{tabular}{l||ccc||cc}
\hline
 & \multicolumn{3}{c||}{Jester} & \multicolumn{2}{|c}{Kinetics-400} \\  \cline{2-4} \cline{5-6} Approach &1\% & 5\% & 10\% & 1\% & 5\% \\ 
\cline{1-1} \cline{2-4} \cline{5-6} 
Supervised (8f) & 52.55$\pm 4.36$ & 85.22$\pm 0.61$ & 90.45$\pm 0.33$ & 6.17$\pm 0.32$ & 20.50$\pm 0.23$ \\
Pseudo-Label~\cite{Lee2013Pseudo} \small{(\texttt{ICMLW'13})} & 57.99$\pm 3.70$ & 87.47$\pm 0.64$ & 90.96$\pm 0.48$ & 6.32$\pm 0.19$ &20.81$\pm 0.86$\\
Mean Teacher~\cite{Tarvainen2017Mean} \small{(\texttt{NeurIPS'17})} & 56.68$\pm 1.46$ & 88.80$\pm 0.44$ & 92.07$\pm 0.03$ & 6.80$\pm 0.42$ & 22.98$\pm 0.43$ \\
S4L~\cite{Zhai2019S4L} \small{(\texttt{ICCV'19})} & 64.98$\pm 2.70$ & 87.23$\pm 0.15$ & 90.81$\pm 0.32$ & 6.32$\pm 0.38$ & 23.33$\pm 0.89$ \\
MixMatch~\cite{Berthelot2019Mixmatch} \small{(\texttt{NeurIPS'19})} & 58.46$\pm 3.26$ & 89.09$\pm 0.21$ & 92.06$\pm 0.46$ & 6.97$\pm 0.48$ & 21.89$\pm 0.22$ \\ 
FixMatch~\cite{Sohn2020Fixmatch} \small{(\texttt{NeurIPS'20})} & 61.50$\pm 0.77$ & 90.20$\pm 0.35$ & 92.62$\pm 0.60$ & 6.38$\pm 0.38$ & 25.65$\pm 0.28$   \\
\hline 
\ours (Ours) & 75.21$\pm 4.48$ & 93.29$\pm 0.24$ & 94.64$\pm 0.21$ &7.69$\pm 0.21$ & 30.28$\pm 0.13$ \\
\ours w/ Finetuning & 77.25$\pm 4.02$ & 93.53$\pm 0.15$ & 94.74$\pm 0.25$ & 8.45$\pm 0.25$ & 31.50$\pm 0.23$ \\
\ours w/ Pretraining \& Finetuning & 82.55$\pm 1.94$ & 93.73$\pm 0.25$ & 94.93$\pm 0.02$ & 11.56$\pm 0.22$ & 31.91$\pm 0.46$\\ 
\hline
\end{tabular}
} \vspace{-2mm}
\caption{\textbf{Performance Comparison on Jester and Kinetics-400}. Numbers show the top-1 accuracy values using ResNet-18 on both datasets. Our approach \ours achieves the best performance across different percentages of labeled data.}
\label{tab:comp_jes_kin} \vspace{-2mm}
\end{table*}

\vspace{1mm}
\noindent\textbf{Baselines.} We compare our approach with the following baselines and existing semi-supervised approaches from 2D image domain extended to video data. First, we consider a supervised baseline where we train an action classifier having the same architecture as the base pathway of our approach.
This is trained using a small portion of the labeled examples assuming only a small subset of labeled examples is available as annotated data.
Second, we compare with state-of-the-art semi-supervised learning approaches, including Pseudo-Label~\cite{Lee2013Pseudo} (ICMLW'13), Mean Teacher~\cite{Tarvainen2017Mean} (NeurIPS'17), S4L~\cite{Zhai2019S4L} (ICCV'19), MixMatch~\cite{Berthelot2019Mixmatch} (NeurIPS'19), and FixMatch~\cite{Sohn2020Fixmatch} (NeurIPS'20). We use same backbone and experimental settings for all the baselines (including our approach) for a fair comparison.

\vspace{1mm}
\noindent\textbf{Implementation Details.} We use Temporal Shift Module (TSM)~\cite{Lin2019Tsm} with ResNet-18 backbone as the base action classifier in all our experiments.
We further investigate performance of different methods by using ResNet-50 on Mini-Something-V2 dataset.
TSM has recently shown to be very effective due to its hardware efficiency and lesser computational complexity.
We use uniformly sampled 8 and 4 frame segments from unlabeled videos as input to the base and the auxiliary pathways respectively to process unlabeled videos in our \ours framework.
On the other hand, we use only 8 frame segments for labeled videos and compute the final performance using 8 frame segments in the base pathway for all the methods.
Note that our approach is agnostic to the backbone architecture and particular values of frame rates.   
Following the standard practice~\cite{Sohn2020Fixmatch} in SSL, we randomly choose a certain percentage of labeled samples as a small labeled set and discard the labels for the remaining data to form a large unlabeled set.
Our approach is trained with different percentages of labeled samples for each dataset (1\%, 5\% and 10\%).
We train our models for 400 epochs where we first train our model with supervised loss $\mathcal{L}_{sup}$ using only labeled data for 50 epochs. We then train our model using the combined loss (ref. Eq.~\ref{eq:total_loss}) for the next 300 epochs. Finally, for finetuning with pseudo-labels, we train our model with both labeled and unlabeled videos having pseudo-label confidence more than 0.8 for 50 epochs.

During pretraining, we follow the standard practice in self-supervised learning~\cite{Chen2020Simple, yang2020video} and train our model using all the training videos without any labels for 200 epochs.
We use SGD~\cite{bottou2010large} with a learning rate of 0.02 and momentum value of 0.9 with cosine learning rate decay in all our experiments. 
Given a mini-batch of labeled samples $B_l$, we utilize $\mu \!\times\! B_l$ unlabeled samples for training.
We set $\mu$ to 3 and $\tau$ to 0.5 in all our experiments. $\gamma$ and $\beta$ values are taken to be 9 and 1 respectively, unless otherwise mentioned.
Random scaling and cropping are used as data augmentation during training (and we further adopt random flipping for Kinetics-400), as in~\cite{Lin2019Tsm}. Following~\cite{Lin2019Tsm}, we use just 1 clip per video and the center 224$\times$224 crop for evaluation.
More implementation details are provided in the appendix.

\subsection{Large-scale Experiments and Comparisons}
\label{subsec:results_analysis}

Tables~\ref{tab:comp_sth_v2_mini}-~\ref{tab:comp_charades} show performance of different methods on all four datasets, in terms of average top-1 clip accuracy and standard deviation over 3 random trials.

\begin{figure}[t!]
  \centering
   \includegraphics[width=\columnwidth]{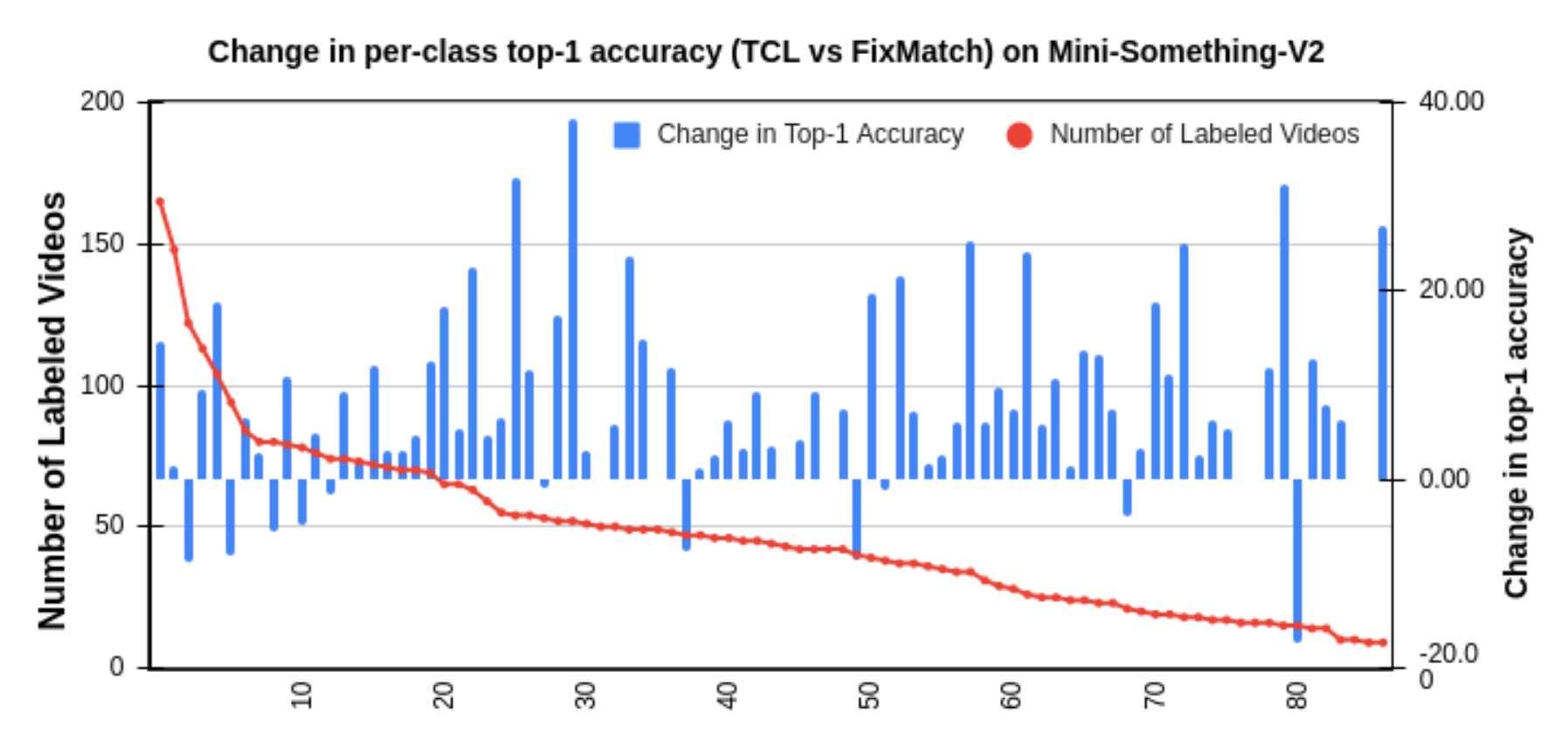} \vspace{-7mm}
   \caption{\small \textbf{Change in classwise top-1 accuracy of \ours over FixMatch on Mini-Something-V2.} Blue bars show the change in accuracy on 5\% labeled scenario, while the red line shows the number of labeled videos per class (sorted). Compared to FixMatch, \ours improves the performance of most classes including those with less labeled data. (Best viewed in color.)}
  \label{fig:class_accuracy} \vspace{-3mm}
\end{figure}

\vspace{1mm}
\noindent\textbf{Mini-Something-V2.}
Table~\ref{tab:comp_sth_v2_mini} shows the performance comparison with both ResNet-18 (left half) and ResNet-50 (right half) backbones on Mini-Something-V2.
\ours outperforms the video extensions of all the semi-supervised image-domain baselines for all three percentages of labeled training data.
The improvement is especially prominent for low capacity model (ResNet-18) and low data (only 1\% and 5\% data with labels) regime. Notably, our approach outperforms the most recent approach, FixMatch by 1.75\% while training with only 1\% labeled data. The improvement is \textbf{8.14\%} for the case when 5\% data is labeled.
These improvements clearly show that our approach is able to leverage the temporal information more effectively compared to FixMatch that focuses on only spatial image augmentations.
Figure~\ref{fig:class_accuracy} shows the classwise improvement over FixMatch along with the number of labeled training data per class in the case of 5\% labeling. The plot shows that a overwhelming majority of the activities experienced improvement with decrease in performance for only 1 class out of 18 having less than 20 labeled videos per class (right of the figure). For low labeled-data regime (1\% and 5\%), a heavier ResNet-50 model shows signs of overfitting as is shown by slight drop in performance. On the other hand, using ResNet-50 backbone instead of ResNet-18 is shown to benefit \ours if the model is fed with more labeled data. Moreover, \ours with finetuning and pretraining shows further improvement, leading to best performance in both cases.

\noindent\textbf{Jester.} Our approach \ours also surpasses the performance of existing semi-supervised approaches in Jester as shown in Table~\ref{tab:comp_jes_kin} (left).
In particular, \ours achieves $10.23\%$ absolute improvement compared to S4L (the next best) in very low labeled-data regime ($1\%$ only).
Adding finetuning and self-supervised pretraining further increases this difference to $17.57\%$. Furthermore, \ours with pretraining and finetuning achieves a top-1 accuracy of 94.93\% using 10\% labeled data which is only 0.32\% lower than the fully supervised baseline trained using all the labels (95.25\%). 

\vspace{1mm}
\noindent\textbf{Kinetics-400.} Table~\ref{tab:comp_jes_kin} (right) summarizes the results on Kinetics-400, which is one of the widely used action recognition datasets consisting of 240K videos across 400 classes.
\ours outperforms FixMatch by a margin of 1.31\% and 4.63\% on 1\% and 5\% scenarios respectively, showing the superiority of our approach on large scale datasets. The top-1 accuracy achieved using \ours with finetuning and pretraining is almost twice better than
the supervised approach when only 1\% of the labeled data is used.
The results also show that off-the-shelf extensions of sophisticated state-of-the-art semi-supervised image classification methods offer little benefit to action classification on videos. 

\vspace{1mm}
\noindent\textbf{Charades-Ego.}
We use third person videos from Charades-Ego~\cite{sigurdsson2018charades} as the target while first person videos form the additional unlabeled set. 
During training, labeled data is taken only from the target domain while unlabeled data is obtained from both the target and the domain-shifted videos.
To modulate domain shift in unlabeled data, we introduce a new hyperparameter $\rho$, whose value denotes the proportion of target videos in the unlabeled set.
For a fixed number of unlabeled videos $|\mathcal{D}_u|$, we randomly select $\rho \!\times\! |D_u|$ videos from the target while the remaining $(1-\rho) \!\times\! |D_u|$ are selected from the other domain.
Following the standard practice~\cite{choi2020unsupervised} in this dataset, we first pretrain the model using Charades~\cite{Sigurdsson2016Hollywood} and experimented using three different values of $\rho$: $1, 0.5, 0$ for 10\% target data with labels.
Table~\ref{tab:comp_charades} shows the mean Average Precision (mAP) of our method including the supervised approach, PseudoLabel and FixMatch. 
\ours outperforms both methods by around 1\% mAP for all three $\rho$ values.
In the case when all the unlabeled data is from the shifted domain ($\rho \!=\! 0$), the performance of our approach is even better than the performance of the next best approach (FixMatch) with $\rho = 1$ \textit{i.e.}, when all unlabeled data is from the target domain itself. This depicts the robustness of \ours and its ability to harness diverse domain data more efficiently in semi-supervised setting.    

\begin{table}[t]
\centering
\resizebox{\columnwidth}{!}{%
\begin{tabular}{l|ccc}
\hline
Approach & \multicolumn{3}{c}{10\%} \\
\hline
Supervised (8f) & \multicolumn{3}{c}{17.53 $\pm{0.49}$}\\
\hline
 & $\rho =$ 1 & $\rho = $0.5 & $\rho = $0\\ 
\cline{2-4}
Pseudo-Label~\cite{Lee2013Pseudo} \small{(\texttt{ICMLW'13})} & 18.00$\pm{0.16}$ &	17.87$\pm{0.14}$ & 17.79$\pm{0.33}$ \\
FixMatch~\cite{Sohn2020Fixmatch} \small{(\texttt{NeurIPS'20})} & 18.02$\pm{0.31}$ &	18.00$\pm{0.29}$ & 17.96$\pm{0.25}$\\
\hline
\ours (Ours) & 19.13$\pm{0.37}$ &	18.95$\pm{0.17}$ & 18.50$\pm{0.95}$\\
\ours w/ Finetuning & 19.68$\pm{0.37}$ &	19.58$\pm{0.31}$&	19.56$\pm{0.82}$ \\
\hline
\end{tabular}
 } \vspace{-1mm}
 \caption{\small \textbf{Semi-supervised action recognition under domain shift (Charades-Ego).} Numbers show mean average precision (mAP) with ResNet-18 backbone across three different proportions of unlabeled data ($\rho$) between third and first person videos. \ours achieves the best mAP, even on this challenging dataset.
 }
 \label{tab:comp_charades} \vspace{-1mm}
 \end{table}

\begin{figure}
    \centering
    \includegraphics[width=\columnwidth]{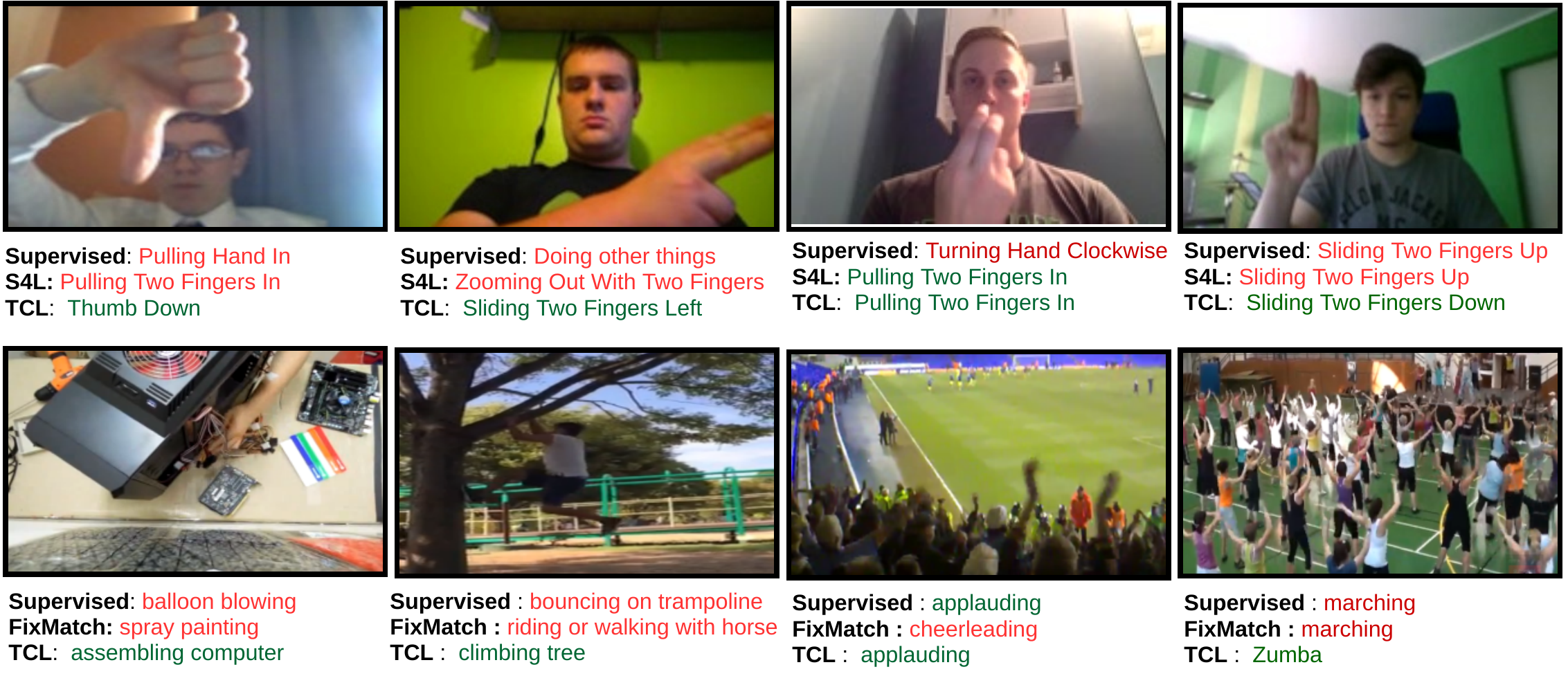} \vspace{-5mm}
    \caption{\small \textbf{Qualitative examples comparing \ours with supervised baseline, S4L~\cite{Zhai2019S4L} and FixMatch~\cite{Sohn2020Fixmatch}.} Top Row: Top-1 predictions using ResNet-18 trained with 1\% labeled data from Jester, Bottom Row: Top-1 predictions using ResNet-18 trained with 5\% labeled data from Kinetics-400. 
    \ours is able to correctly recognize different hand gestures in Jester and diverse human actions in Kinetics-400 dataset. (Best viewed in color.)
    }
    \label{fig:qualitative_eample} \vspace{-3mm}
\end{figure}

\vspace{1mm}
\noindent\textbf{Qualitative Results.} 
Figure~\ref{fig:qualitative_eample} shows qualitative comparison between our approach \ours and other competing methods (S4L~\cite{Zhai2019S4L} and FixMatch~\cite{Sohn2020Fixmatch}) including the simple supervised baseline on Jester and Kinetics-400 respectively.
As can be seen, our temporal contrastive learning approach is able to correctly recognize different hand gestures from Jester dataset even with 1\% of labeling, while the supervised baseline and the next best approach (S4L) fail to recognize such actions. Similarly, our approach by effectively utilizing temporal information, predicts the correct label in most cases including challenging actions like `climbing tree' and `zumba' on Kinetics-400 dataset. More qualitative examples are included in the appendix. 

\vspace{1mm}
\noindent\textbf{Role of Pseudo-Labeling.} We test the reliability of pseudo-labeling on Jester (using ResNet-18 and 1\% labeling) with 50 epoch intervals and observe that the pseudo-labeling accuracy gradually increases from 0\% at the beginning to 65.95\% at 100 epoch and then 93.23\% at 350 epoch. This shows that while our model may create some wrong groups at the start, it gradually improves the groups
as the training goes by, leading to a better representation by exploiting both instance and group contrastive losses.

\vspace{-3mm}
\subsection{Ablation Studies}

We perform extensive ablation studies on Mini-Something-V2 with 5\% labeled data and ResNet-18 backbone to better understand the effect of different losses and hyperparameters in our framework.

\vspace{1mm}
\noindent\textbf{Effect of Group Contrastive Loss.} We perform an experiment by removing group contrastive loss from our framework (ref. Section~\ref{subsubsec:group-consistency}) and observe that top-1 accuracy drops to 27.24\% from 29.81\% (Table \ref{tab:ablation}), showing the importance of it in capturing high-level semantics.

\vspace{1mm}
\noindent\textbf{Ablation on Contrastive Loss.} We investigate the effectiveness of our contrastive loss by replacing it with pseudo-label consistency loss used in FixMatch~\cite{Sohn2020Fixmatch}.
We observe that training with our contrastive loss, surpasses the performance of the training with the pseudo-label consistency loss by a high margin (around 6.21\% gain in the top-1 accuracy) on Mini-Something-V2 (Table~\ref{tab:ablation}).
We further compare our approach in the absence of group-consistency (\ours w/o Group-Contrastive Loss) with a variant of FixMatch~\cite{Sohn2020Fixmatch} that uses temporal augmentation and observe that our approach still outperforms it by a margin of 2.66\% (24.58\% vs 27.24\%) on Mini-Something-V2 (with ResNet-18 and 5\% labeling). This shows that temporal augmentation alone fails to obtain superior performance and this improvement is in fact due to the efficacy of our contrastive loss formulation over the pseudo-label loss used in FixMatch~\cite{Sohn2020Fixmatch}. 

\vspace{1mm}
\noindent\textbf{Effect of Different Frame Rate.}
We analyze the effect of doubling frame-rates in both pathways and observe that \ours (w/ 16 frame segments in base and 8 frame segments in the auxiliary pathway) improves top-1 accuracy by 1.5\% on Mini-Something-V2 with ResNet-18 and 5\% labeled data (29.81\% vs 31.31\%).
However, due to heavy increase in compute and memory requirement with little relative gain in performance, we limit our study to 8 and 4 frame setting.

\begin{table}[t]
\centering
\resizebox{\columnwidth}{!}{%
\begin{tabular}{l|c}
\hline
Approach &  Top-1 Accuracy  \\
\hline
\ours w/o Group-Contrastive Loss & 27.24$\pm{0.42}$\\
\ours w/ Pseudo-Label Consistency Loss & 23.60$\pm{1.04}$ \\
\hline
\ours (Ours) & 29.81$\pm{0.77}$\\
\hline
\end{tabular}
} \vspace{-1mm}
\caption{\small \textbf{Ablation Studies on Mini-Something-V2.} Numbers show top-1 accuracy with ResNet-18 and 5\% labeled Data.}
\label{tab:ablation} \vspace{-2mm}
\end{table}

\begin{figure}[t]
  \centering
   \includegraphics[width=1\columnwidth]{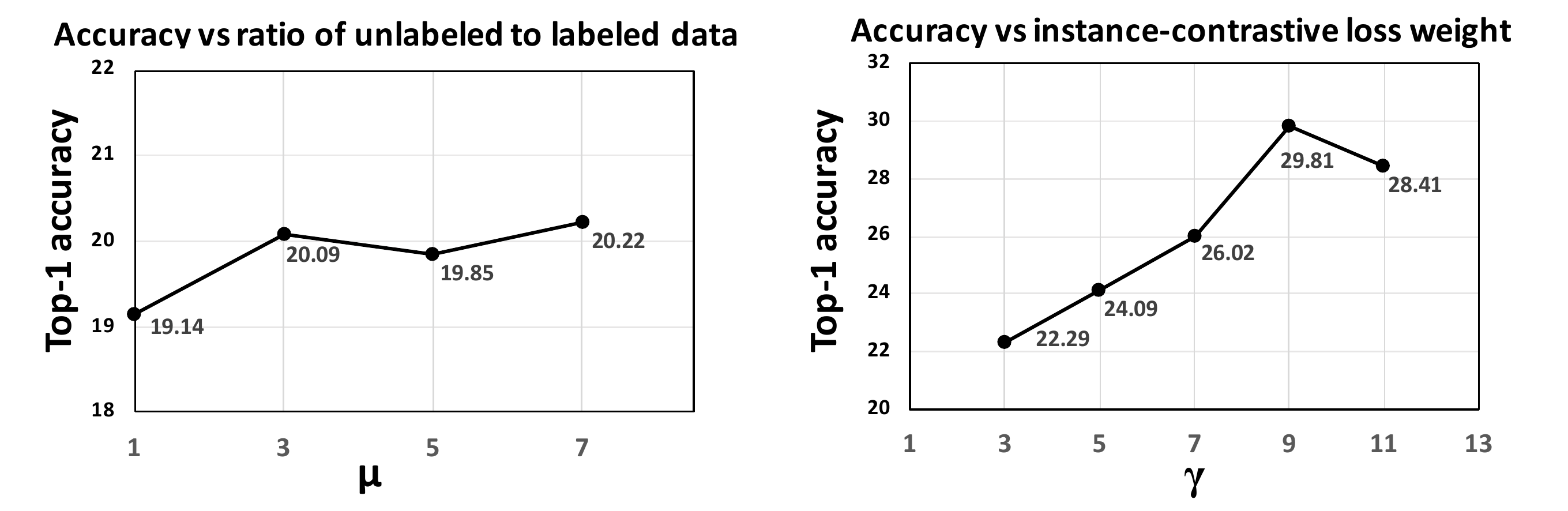} \vspace{-7mm}
  \caption{\small \textbf{Effect of Hyperparameters on Mini-Something-V2.} (Left) Varying the ratio of unlabeled data to the labeled data ($\mu$), (Right) Varying the instance-contrastive loss weight ($\gamma$).
  }
  \label{fig:hyper} \vspace{-2mm}
\end{figure}

\vspace{1mm}
\noindent\textbf{Effect of Hyperparameters.} We analyze the effect of the ratio of unlabeled  data to labeled data ($\mu$) and observe that by setting $\mu$ to \{3, 5, 7\} with a fixed $\gamma=1$, produces similar results on Mini-Something-V2 (Figure~\ref{fig:hyper} (Left)). However, as scaling $\mu$ often requires high computational resources, we set it to 3 in all our experiments to balance the efficiency and accuracy in semi-supervised action recognition. We also find that weight of the instance-contrastive loss ($\gamma$) greatly affects the performance in semi-supervised learning as accuracy drops by more than 6\% when setting $\gamma$ to 3 instead of the optimal value of 9 on Mini-Something-V2 with ResNet-18 backbone and 5\% of labeling (Figure~\ref{fig:hyper} (Right)).

\vspace{1mm}

\noindent\textbf{Comparison With Self-Supervised Approaches.} We compare our method with three video self-supervised methods, namely Odd-One-Out Networks (O3N)~\cite{fernando2017self}, Video Clip Order Prediction (COP)~\cite{Xu2019Self} and Memory-augmented Dense Predictive Coding (MemDPC)~\cite{Han2020Memory} through pretraining using self-supervised method and then finetuning using available labels on Mini-Something-V2 (with ResNet18 and 5\% labeled data).
Our approach significantly outperforms all the compared methods by a margin of 6\%-10\% (O3N: 19.56\%, COP: 23.93\%, MemDPC: 18.67\%, \ours: 29.81\%), showing its effectiveness over self-supervised methods.
Moreover, we also replace our temporal contrastive learning with O3N and observe that accuracy drops to 24.58\% from 29.81\%, showing the efficacy of our contrastive learning formulation over the alternate video-based self-supervised method on Mini-Something-V2.

\section{Conclusion}
\label{sec:conclusions}
We present a novel temporal contrastive learning framework for semi-supervised action recognition by maximizing the similarity between encoded representations of the same unlabeled video at two different speeds as well as minimizing the similarity between different unlabeled videos run at different speeds. We employ contrastive loss between different video instances including groups of videos with similar actions to explore high-level action semantics within the neighborhood of different videos depicting different instances of the same action. We demonstrate the effectiveness of our approach on four standard benchmark datasets, significantly outperforming several competing methods.

\vspace{1mm}
{\small \noindent\textbf{Acknowledgements.} This work was partially supported by the SERB Grant SRG/2019/001205. This work is also supported by the Intelligence Advanced Research Projects Activity (IARPA) via DOI/IBC contract number D17PC00341. The U.S. Government is authorized to reproduce and distribute reprints for Governmental purposes notwithstanding any copyright annotation thereon.
The views and conclusions contained herein are those of the authors and should not be interpreted as necessarily representing the official policies or endorsements, either expressed or implied, of IARPA, DOI/IBC, or the U.S. Government.}

\clearpage
{\small
\bibliographystyle{ieee_fullname}
\bibliography{video_ssl}
}

\newpage
\appendix
\section*{\centering{Appendix}}
The appendix contains the following.
\begin{itemize}
\setlength{\itemsep}{-0.2pt}
\item Section~\ref{sec:dataset}: Dataset details used in our experiments.
\item Section~\ref{sec:implementation}: Implementation details of our \ours framework.
\item Section~\ref{sec:bl}: Implementation details of the video extensions of the image-based baselines.
\item Section~\ref{sec:acc}: Additional classwise improvements over S4L for  $1\%$ labeled data in Jester.
\item Section \ref{sec:group}: Effect of group contrastive loss on image datasets.
\item Section~\ref{sec:qeg}: Additional qualitative examples from different datasets.
\end{itemize}

\section{Dataset-Details}
\label{sec:dataset}

\vspace{1mm}
\noindent\textbf{Mini-Something-V2.} The Mini-Something-V2 dataset~\cite{chen2020deep} is a subset of Something-Something V2 dataset~\cite{Goyal2017Something}. It contains a total of $81663$ training videos and $11799$ validation videos. The resolution of each video is set to a height of $240$px and has an average duration of $4.03$ seconds. There are a total of $87$ action classes related to basic object interactions such as `Putting something into something', `Showing something behind something', `Squeezing something' and `Showing that something is inside something'.

\vspace{1mm}
\noindent\textbf{Jester.} The jester dataset consists of a total of 148,092 videos spread across 27 classes with an average of 4391 per class samples. 
The classes belong to a series of hand gestures such as `Sliding Two Fingers Up', `Turning Hand Clockwise' and `Swiping Down'. 
Specifically, the training set contains a total of 118,562 clips and 14,787 clips are provided for validation. The average duration of the videos are 3 seconds.
The frames are extracted from these videos with 12 fps and maintain a fixed height of 100px but with variable width. The dataset is publicly available at \href{https://20bn.com/datasets/jester/v1}{https://20bn.com/datasets/jester/v1}.

\vspace{1mm}
\noindent\textbf{Kinetics-400.} The Kinetics-400 is a benchmark dataset containing YouTube videos of diverse human-action classes. It consists of around $300$K videos spread across $400$ classes with each class containing atleast $400$ clips. The classes range across a broad spectrum of actions such as shaking hands, hugging and playing instruments. This dataset can be obtained from the link, \href{https://deepmind.com/research/open-source/kinetics}{https://deepmind.com/research/open-source/kinetics}.

\vspace{1mm}
\noindent\textbf{Charades-Ego.} The Charades-Ego dataset is one of the largest datasets comprising of both first-person and third-person views of videos collected across a diverse set of $112$ actors. 
The total $7,860$ samples consist of around $4000$ such pairs, each spanning around $31.2$ seconds on average at $24$ fps. The videos in this dataset have multiple activity classes which often overlap, making the dataset particularly challenging. The training set is divided into two separate lists, `CharadesEgo\_v1\_train\_only3rd' and `CharadesEgo\_v1\_train\_only1st', which contain the videos corresonding to the third-person and first-person perespectives respectively. Each file lists the video ids with their corresponding activity classes.
Following the standard practice~\cite{zhou2018temporal}, we first trim the multi-class 3082 videos of `CharadesEgo\_v1\_train\_only3rd' and 3085 videos of `CharadesEgo\_v1\_train\_only1st' to obtain 34254 and 33081 single-class clips respectively. We select the 10\% labeled videos class-wise from the 34254 trimmed clips distributed over 157 activity classes.
The mAP metric is evaluated over the full `CharadesEgo\_v1\_test\_only3rd' video set.
The dataset is publicly available at 
\href{https://github.com/gsig/actor-observer}{https://github.com/gsig/actor-observer}.

\section{Implementation Details}
\label{sec:implementation}
In this section, we provide additional implementation details (refer Section 4.1 of the main paper) of our \ours framework. For the basic convolution operation over the videos, we use the approach identical to that of Temporal Segment Network (TSM)~\cite{Lin2019Tsm}. We utilize the 2D CNNs for their lesser computational complexity over the 3D counterparts and implement the bi-directional temporal shift module to move the feature channels along the temporal dimension to capture the temporal modeling of the samples efficiently. All hyperparameters related to TSM module has been taken from ~\cite{Lin2019Tsm}. As shown in ~\cite{Lin2019Tsm}, this approach achieves the state-of-art performances while significantly reducing the computational complexity. We have considered 2D ResNet-18 model as our primary backbone and have incorporated the temporal shift module after every residual branch of the model to avoid the interference with the spatial feature learning capability.
In our experiments, one epoch has been defined as one pass through all the labeled data.
We have used learning rate of $0.002$ during the finetuning stage.

\section{Image-based Baseline Details}
\label{sec:bl}
This section provides implementations details of different baselines used in the paper. We have adhered to the base approach proposed in the original works of the respective baselines for all our experiments.
Note that, for a given video, same set of augmentations have been applied to all frames of the video so that all frames in a video go through the same set of transformations and do not loose the temporal consistency between the them.
Also, following TSM~\cite{Lin2019Tsm}, for the high spatially-sensitive datasets like Mini-Something-V2~\cite{Goyal2017Something} and Jester~\cite{materzynska2019jester}, we refrain from using the \textit{Random Horizontal Flip} as it may effect the spatial semantics of the frames.
The initial lr is set to $0.02$ with cosine learning decay in all our baseline experiments unless stated otherwise. All the baselines models are trained for $350$ epochs unless otherwise specified.

\vspace{1mm}
\noindent\textbf{Supervised}
We have used the code made public by the authors in~\cite{Lin2019Tsm} for the supervised baseline.
It is trained using $\mathcal{L}_{sup}$ for 200 epochs and the initial learning rate is kept same as in TCL.
Other hyperparameters are kept same as the ones used for the respective datasets in~\cite{Lin2019Tsm}.

\vspace{1mm}
\noindent\textbf{MixMatch} We followed the approach in ~\cite{Berthelot2019Mixmatch} to train our MixMatch baseline approach.
We applied $2$ different augmentations to unlabeled videos set ($U$) and then computed the average of the predictions across these augmentations.
We have used cropping and flipping as the two augmentations in our experiments.
The sharpened versions of the average predictions of $K$ different augmentations are used as labels for the unlabeled videos.
Then, labeled ($V$) and unlabeled videos with their targets and predicted labels are shuffled and concatenated to form another set $W$ which serves as a source for modified MixUp algorithm defined in ~\cite{Berthelot2019Mixmatch}.
Then for each $i^{th}$ labeled video we compute \textit{MixUp}$(V_i,W_i)$ and add the result to a set $V'$. It contains the \textit{MixUp} of labeled videos with $W$.
Similarly for each $j^{th}$ unlabeled video, we compute \textit{MixUp}$(U_i,W_{i+|V|})$ and add the result to another set  $U'$. It contains the \textit{MixUp} of unlabeled videos with rest of $W$.
A cross-entropy loss between labels and model predictions from $V'$ and MSE loss between the predictions and guessed labels from $U'$ are used for training.
The temperature is set to $0.5$ and both $\mu$ and $\gamma$ are set to $1$.

\vspace{1mm}
\noindent\textbf{S4L}: S4L~\cite{Zhai2019S4L} is a self-supervised semi-supervised baseline used in our work.
The self-supervision is done by rotating the input videos.
Videos are rotated by $\{ 0, 90, 180, 270\}$ degrees and the model is trained to predict these rotations of the videos.
The corresponding rotation loss~\cite{Zhai2019S4L} is used for both labeled and unlabeled videos.
The $\mu$ and $\gamma$ are set to $5$ in this baseline experiment.
The S4L model is trained using rotation loss apart from the $\mathcal{L}_{sup}$ for labeled videos.
The initial learning rate is set to $0.1$.

\vspace{1mm}
\noindent\textbf{Pseudo-Label} Pseudo-label~\cite{Lee2013Pseudo} leverages the idea that in absence of huge amount of labeled data, artificial labels or pseudo-labels for unlabeled data should be obtained using the model itself.
Following this basic intuition, we first train our model using $\mathcal{L}_{sup}$ for 50 epochs to get a reasonably trained model.
The next $300$ epochs are run using both labeled and unlabeled videos.
Consistency is ensured between the pseudo-labels of the unlabeled video with the logits predicted for them by the model.
The class for which an unlabeled video gets the highest activation from the model is taken as the pseudo-label of it.
Only videos which have highest activation greater than $0.95$ are assigned pseudo-labels.
Both $\mu$ and $\gamma$ are set to $3$ in this set of experiments.

\vspace{1mm}
\noindent\textbf{MeanTeacher} : The model is trained using the philosophy described in~\cite{Tarvainen2017Mean}.
In this scenario, we have two models, one is the \emph{student} network and the other is the \emph{teacher} network. The teacher network has the same backbone architecture as the student.
The weights of the teacher network are exponential moving average weights of the student network.
Consistency is ensured between the logits predicted by the teacher and the student for the unlabeled videos. The labeled data, in addition, is trained using $\mathcal{L}_{sup}$. Both $\mu$ and $\gamma$ are set to $1$ in this set of experiments. $\gamma$ is increased from $0$ to $1$ using sigmoid function over 50 epochs as in ~\cite{Tarvainen2017Mean}.

\vspace{1mm}
\noindent\textbf{FixMatch.} For extending the FixMatch baseline to video domain, we primarily follow the same augmentation and consistency regularization policies laid out in ~\cite{Sohn2020Fixmatch}. The videos are passed through two different pathways. In the first pathway, the video frames are weakly augmented and used to obtain the pseudo-labels. In the second pathway, the strongly augmented versions of the same video frames are trained for their representations to be consistent with the corresponding pseudo-labels.
Specifically, in the case of weak augmentations, we use \textit{Random Horizontal Flip} followed by \textit{Random Vertical and Horizontal shifts}. For the strong augmentations we use the \textit{RandAugment}~\cite{cubuk2020randaugment} augmentation policy followed by \textit{CutOut} augmentation. The experiments are carried out for 350 epochs with a batch size of 8 and considering the $\mu$ and $\gamma$ values as 3 and 9 respectively.

\begin{figure}
	\centering
	\includegraphics[width=\columnwidth]{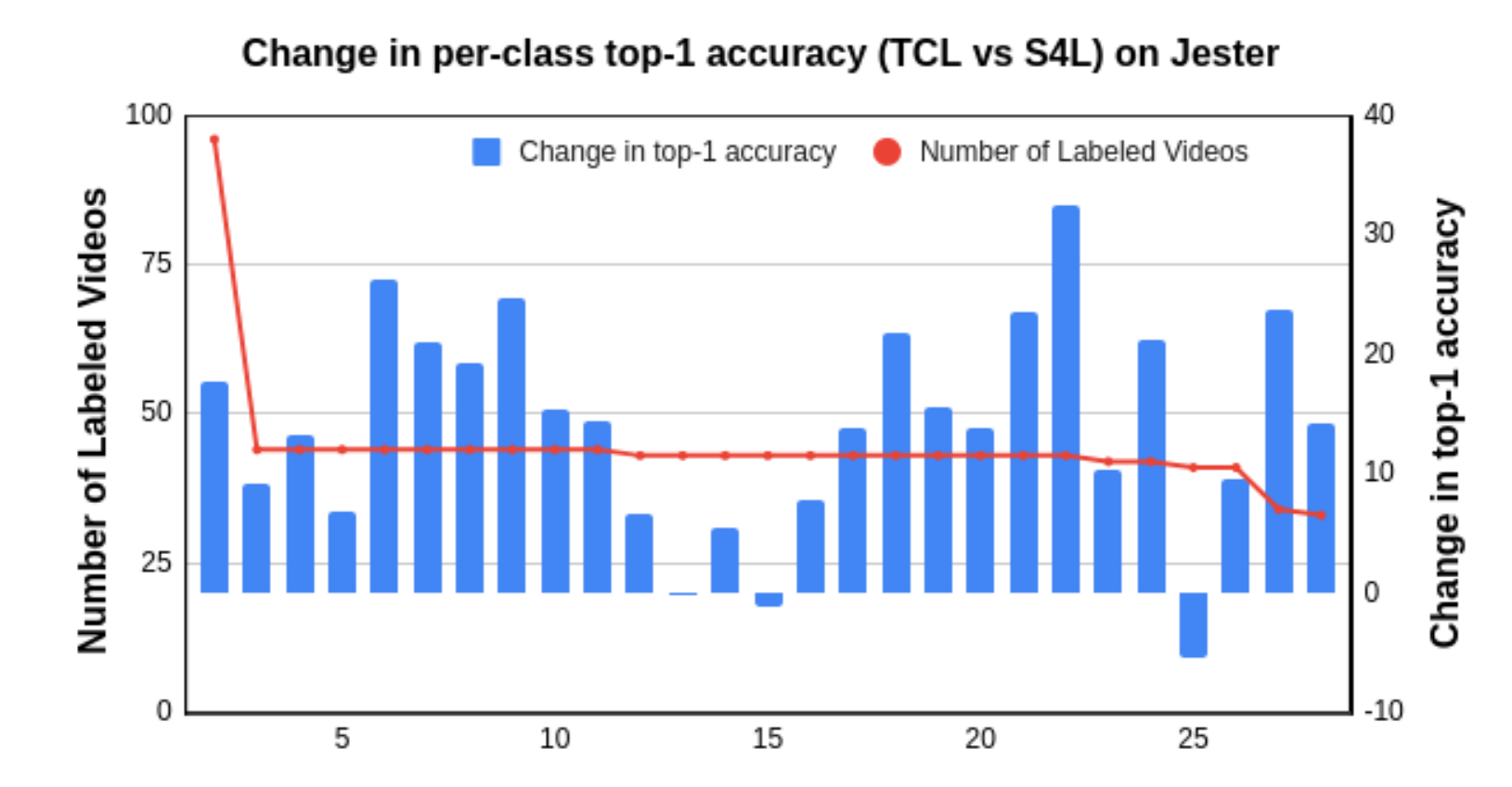}
	\caption{\small \textbf{Change in classwise top-1 accuracy of \ours over S4L on Jester.} {\color{blue}Blue} bars show the change in accuracy on 1\% labeled scenario of Jester dataset. The {\color{red}red} line depicts the number of labeled videos per class in a sorted manner. Compared to S4L, \ours improves the performance of most classes including those with less labeled data. (Best viewed in color.)}
	\label{fig:class_accuracy_jester}
\end{figure}

\section{Classwise Improvements}
\label{sec:acc}

In the main paper, we have presented the change in top-1 accuracy per class of \ours over FixMatch on $5\%$ Mini-Something V2. Here, we have included the change in top-1 accuracy per class of \ours over S4L (next best) on Jester dataset using only $1\%$ labeled data in Figure \ref{fig:class_accuracy_jester}.
We can observe in Figure~\ref{fig:class_accuracy_jester} that only $2$ classes in Jester have less improvement over S4L for this $1\%$ labeled data scenario.

\section{Group Contrastive Loss on Image Dataset} 
\label{sec:group}
We analyze the effect of group contrastive loss on CIFAR10 (using SimCLR~\cite{Chen2020Simple} with WideResNet-28-2 and 4 labeled samples per class) and observe that it improves performance by $3.15\%$ ($84.11\%$ vs $87.26\%$), showing the effectiveness of group contrastive loss in semi-supervised classification on image datasets too besides the video datasets.

\section{Qualitative Examples}
\label{sec:qeg}

\begin{figure*}[t]
    \centering
    \includegraphics[width=2\columnwidth]{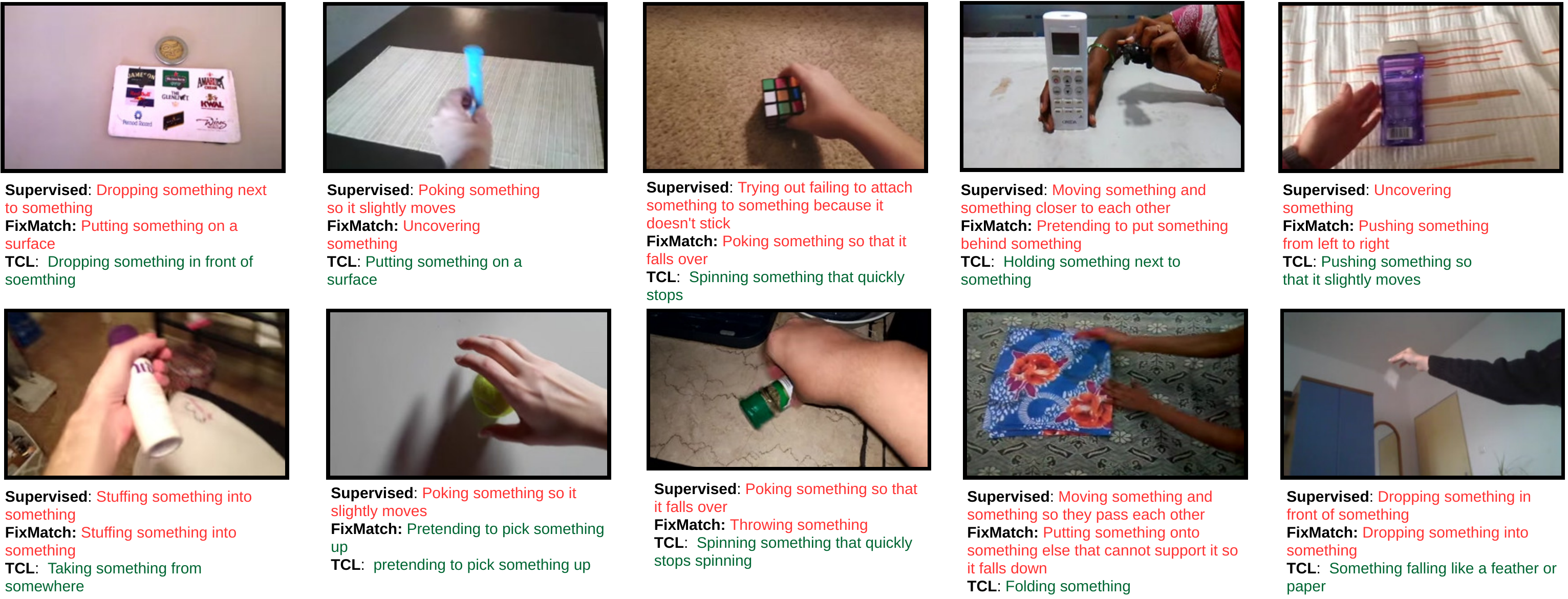}
    \caption{\small \textbf{Qualitative examples comparing \ours with supervised baseline and FixMatch ~\cite{Sohn2020Fixmatch} on Mini-Something V2 trained using $5\%$ labeled data with ResNet-18}. Both rows provide top-1 predictions using supervised baseline, FixMatch and proposed \ours approach respectively from top to bottom. As observed, the supervised baseline trained using only the labeled data predicts wrong actions. While the competing methods fail to classify the correct actions in most cases \ours is able to correctly recognize different actions in this dataset. The predictions marked in {\textcolor{ForestGreen}{green}} match the ground truth labels, whereas the {\color{red}{red}} marked predictions are wrong. (Best viewed in color.)}
    \label{fig:samples_something}
\end{figure*}

\begin{figure*}[h]
    \centering
    \includegraphics[width=2\columnwidth]{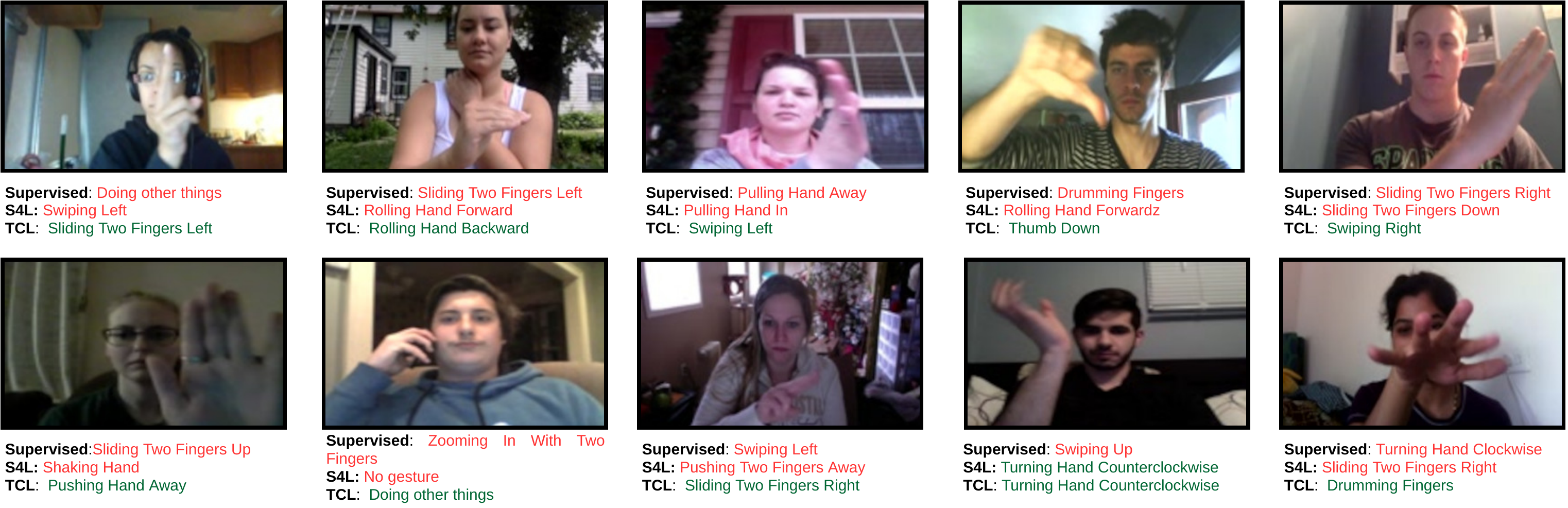}
    \caption{\small \textbf{Qualitative examples comparing \ours with supervised baseline and S4L~\cite{Zhai2019S4L} on Jester dataset trained using 1\% labeled data with ResNet-18.} Both rows provide top-1 predictions using supervised baseline, S4L and \ours respectively from top to bottom. As observed, the supervised baseline trained using only the labeled data predicts wrong actions. While the competing methods fail to classify the correct actions in most cases, our proposed approach, \ours is able to correctly recognize different hand gestures in this dataset. The predictions marked in {\textcolor{ForestGreen}{green}} match the ground truth labels, whereas the {\color{red}{red}} marked predictions are wrong. (Best viewed in color.)}
    \label{fig:samples_jester}`
\end{figure*}

\begin{figure*}[h]
    \centering
    \includegraphics[width=2\columnwidth]{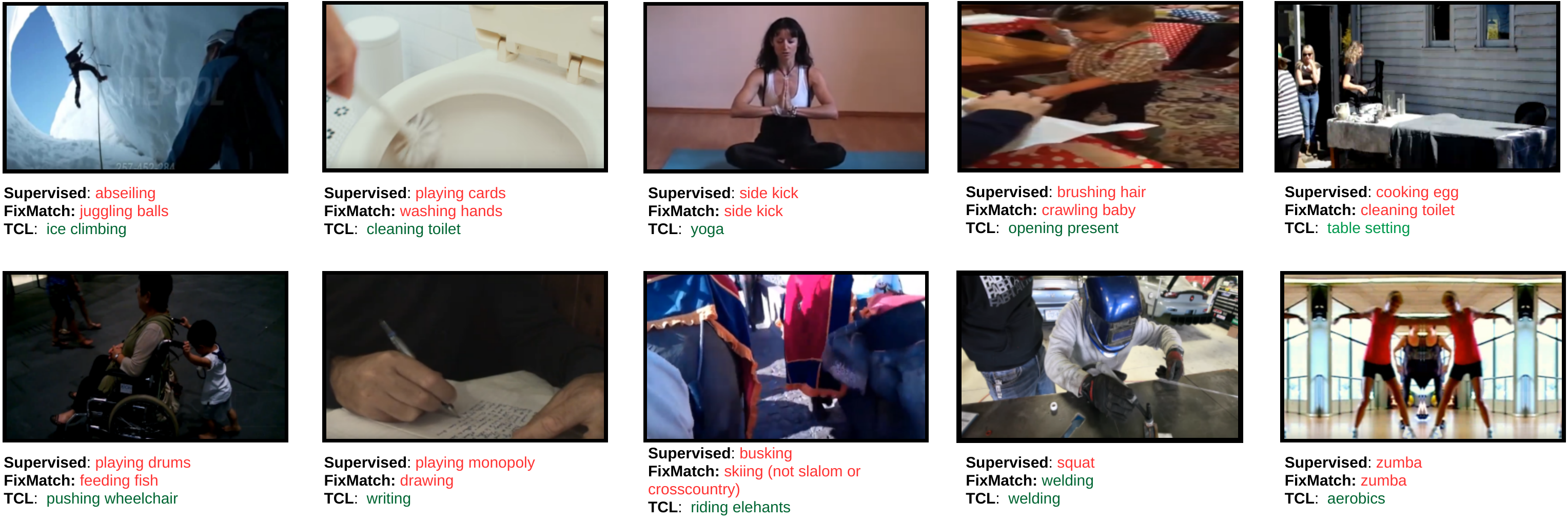}
    \caption{\small \textbf{Qualitative examples comparing \ours with supervised baseline and FixMatch~\cite{Sohn2020Fixmatch} on Kinetics-400 trained using $5\%$ labeled data with ResNet-18}. Both rows provide top-1 predictions using supervised baseline, FixMatch and \ours respectively from top to bottom. As observed, the supervised baseline trained using only the labeled data predicts wrong actions. While the competing methods fail to classify the correct actions in most cases our proposed approach, \ours is able to correctly recognize different actions in this dataset. The predictions marked in {\textcolor{ForestGreen}{green}} match the ground truth labels, whereas the {\color{red}{red}} marked predictions are wrong. (Best viewed in color.)}
    \label{fig:samples_kinetics}
\end{figure*}

\begin{figure*}[h]
    \centering
    \includegraphics[width=2\columnwidth]{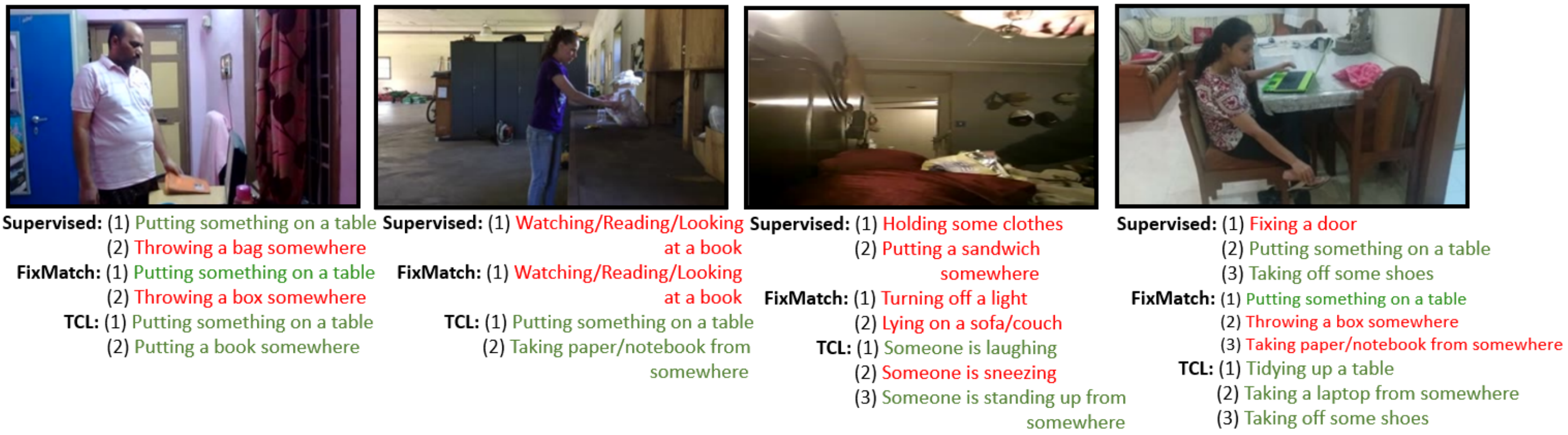} 
    \caption{\small \textbf{Qualitative examples comparing \ours with supervised baseline and FixMatch~\cite{Sohn2020Fixmatch} on Charades-Ego}. As each of the video samples have multiple actions, we show random frames from different videos of the dataset and compare the Top-K predictions for those frames. Here, `K' denotes the number of ground-truth classes associated with the respective samples.  While the supervised and competing methods fail to classify all the correct actions in most cases, \ours is able to correctly recognize most of the relevant actions in these videos. The predictions marked in {\textcolor{ForestGreen}{green}} match ground truth labels, whereas {\color{red}{red}} marked predictions are wrong. (Best viewed in color.)
    }
    \label{fig:charadesego_quality} 
\end{figure*}

In the Main paper, we provided qualitative examples from Jester and kinetics-400 dataset. Here we have included some more samples from all four datasets to show the superiority of our methods over the competing baseline methods.
Figure \ref{fig:samples_something}, \ref{fig:samples_jester}, \ref{fig:samples_kinetics} and \ref{fig:charadesego_quality} contain the example frames and their predictions for Mini-Something V2, Jester, Kinetics-400 and Charades-ego respectively.

\end{document}